# Comparison between ARIMA and Deep Learning Models for Temperature Forecasting

**Eranga De Saa**
*Department of information Technology*
*University of Moratuwa*
Katubedda, Sri Lanka
erangadesaa@gmail.com

**Lochandaka Ranathunga**
*Department of information Technology*
*University of Moratuwa*
Katubedda, Sri Lanka
lochandaka@uom.lk

*Abstract*- **Weather forecasting benefits us in various ways from farmers in cultivation and harvesting their crops to airlines to schedule their flights. Weather forecasting is a challenging task due to the chaotic nature of the atmosphere. Therefore lot of research attention has drawn to obtain the benefits and to overcome the challenges of weather forecasting. This paper compares ARIMA (Auto Regressive Integrated Moving Average) model and deep learning models to forecast temperature. The deep learning model consists of one dimensional convolutional layers to extract spatial features and LSTM layers to extract temporal features. Both of these models are applied to hourly temperature data set from Szeged, Hungry. According to the experimental results deep learning model was able to perform better than the traditional ARIMA methodology.**

I. INTRODUCTION

Weather forecasting is essential for us to know unforeseeable information that will aid us when carrying daily tasks. For example farmers could use this futuristic information to cultivate crops on time. Further, Air lines could schedule flights safely and accurately. Other than that, these predictions could use to notify people when imminent dangers such as tsunamis, hurricanes are near. This information helps us to make important daily decisions. Weather forecasting is a challenging task due to the uncertain, chaotic nature of the atmospheric conditions and lack of understanding of the various atmospheric processes. Traditionally weather forecasting was done through using climatology and using analog methods. Climatology applied by getting an average of weather statistics gathered over several years to make the prediction. Analog method predicts weather assuming weather in present will behave the same way as it did behave in past.

Present there are two methods for numerical weather prediction. First method is using weather data. This method includes Support vector machine [1] , regression [2], Naïve bayes based machine learning approaches [3]. Second method is prediction based on time series data. This method includes usage of artificial neural network [4], ARMA [5], ARIMA [6] model based Approaches. This study aims to build forecasting models using ARIMA and deep learning. And comparatively analyzes prediction results of each of these models based on performance. A historical weather data from Szeged, Hungry between 2006 and 2016 time period is obtained from keggle website to model training and validation.

This paper follows the following structure: Section II goes through related literature; Section III describes the data set and preprocessing ; Section IV introduces ARIMA model and proposed deep learning model ; then section V illustrates how the model evaluation is performed ; next Section VI Analyzes the experimental prediction results ; and lastly Section VII discusses the conclusion.

II. RELATED WORK

According to a study conducted by Mark Hallstrom, Dylan Liu and Christopher Vo to exert machine learning to predict weather [2]. The illustrated method in this study uses data collected from Weather Underground. The collected data set includes minimum temperature, maximum temperature and mean pressure of the atmosphere, mean humidity and daily weather condition during the years of 2011 – 2015 in Stanford. This study is constricted to predict minimum and the maximum temperature for seven days ahead by using the data from the past two days. Four weather classifications which are namely rain, precipitation, cloudy and very cloudy were used in this study. Linear regression algorithm was used to forecast maximum and minimum temperature as a linear combination. Further, a variation of functional algorithms was used to identify previous weather conditions that were analogous to the current weather condition. Then these past weather conditions were used to predict the weather.

Moreover, this study has tested functional regression and linear regression models with few days. Further, both professional weather forecasting services were able to outperform both the of machine learning models that were discussed in the study. It was found that the professional weather forecasting services performances decreased over time. As for the future improvements this study proposes to collect more data to increase the performance of the linear regression models.

According to study by Tarun Rao, N. Rajasekhar and T. V Rajinikanth to forecast weather using support vector machines as a resourceful approach [1]. Support vector machines algorithm falls in to supervised machine leaning model category and they're widely used for classification and regression analysis problems. In this research linear support vector regression is utilized to predict maximum temperature of the day. Further, in this study multi layer perceptrons which are trained with back propagation algorithm is used to compare with the support vector machine model to predict weather. Perceptrons which are widely known as neural networks are also mostly falls in to supervised machine learning algorithm category. Neural networks mimic the way animal and humans learn in order to develop a learning model. Back propagation is the method that is used to fine tune the weights of each neuron in the network by backward propagating the error to inner layers of the network. To train each of these models this study has used a data set from University of Cambridge which has a span of 5 years.

According to the results the utilization of support vector machines for weather forecasting was able to outperform the multi layer perceptron neural network which is trained by back propagation algorithm. The only drawback is support vector machines training time is increased when dealing with higher orders.

In a Study done by Lai et al. on weather forecasting using dynamically weighted time delay neural networks in the east china [7]. Data from fourteen local weather canters during 1991 to 2001 time period is used to train and validate the dynamic weighted time delay neural network used in the study. Dynamic weighted time delay neural network is a sub category of neural networks. Time delay neural networks has the ability to identify translation invariant shapes and able to model the contest at every layer.

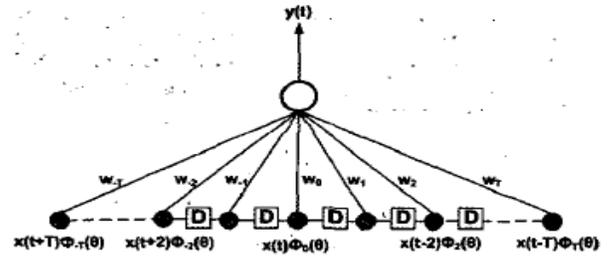

Figure.1 Architecture of a dynamic weighted time-delay neuron [7]

According to the results of this study the author has come to a conclusion that an even neural network with a single hidden layer can approximately predict the rainfall and temperature.

In a study done by Warangkhana Kimpan and Saktaya Suksri describes about neural network model for weather prediction using fireworks algorithm [8]. In this study the neural network is trained via the fireworks algorithm. Fireworks algorithm is a swarm intelligence algorithm which mimics the group behavior. This algorithm is mainly used for optimization purposes. The aim of this study is to forecast the mean temperature of a day. This study uses past daily weather between 2012 and 2015 from Chaloemprakiet weather station as the dataset. The model attained training accuracy of 81.48% and testing accuracy of 73.79%. This study suggests comparison with other prediction algorithms as for the future works

### III. DATASET AND PREPROCESSING

Weather data from Szeged, Hungry has been used for this study. Data is collected from the keggle website. Data set contains hourly weather data such as temperature, humidity and wind speed. Only temperature values were used for this study. Data set covers the time period between 2006 and 2016.

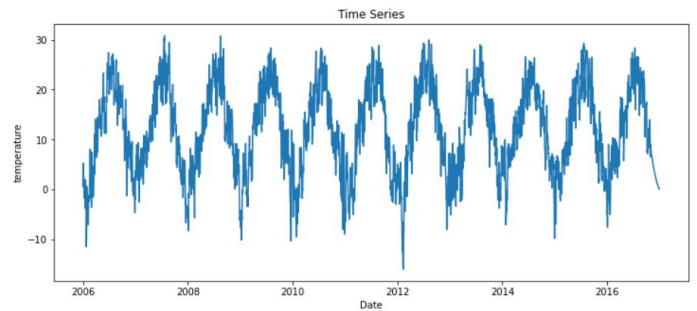

Figure.2 Daily temperature time series

To simplify the analysis only the only monthly mean temperature values were taken for the consideration.

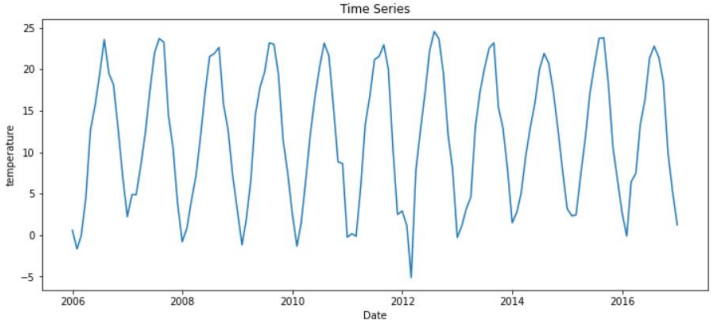

Figure.3 Monthly temperature time series

Data set should be normalized data in order to be trained by a machine learning algorithm. Therefore, temperature values of the time series data set was normalized using z-score normalization technique. Where z score is calculated based on the temperature sample mean ($\bar{x}$) and sample standard deviation ($S$). Thereby, normalized value z can be obtained for the time series data $x_i$ by following equation.

$$Z = \frac{x_i - \bar{x}}{S} \quad (1)$$

## IV. FORECASTING MODELS

### A. ARIMA

According to research conducted by Box and Jenkins [9] ARIMA ((p, d, q) Auto regressive integrated moving average model) which is a part of stochastic processes were used to analyze time series data. Since then ARIMA models have been widely applies in a wide range of time series analysis applications. According to research done by Ling Chen and Xu Lai [6] which involves wind speed forecasting by applying ARIMA models methodology of box and Jenkins as follows,

Step 1 : Model recognition

Step 2 : Estimation of parameters

Step 3 : Checking the residual diagnostics and forecasting

#### 1) Model recognition

ARIMA models are one of the most widely used approaches for time series forecasting. For a time series which is stationary ARIMA (p, d, q) Model can be written in terms of past temperature data, residuals and prediction errors as follows,

$$x_t = \sum_{i=1}^{p} \varphi_i x_{t-i} - \sum_{j=1}^{q} \theta_j a_{t-j} + a_t \quad a_t \sim NID(o, \sigma^2) \quad (2)$$

Temperature time series data is denoted by $x$ and (t-i)th data is denoted by $x_{t-i}$, random white noise time series is denoted by $a_t$, auto regressive parameters are denoted by $\varphi_i$, moving average parameters are denoted by $\theta_j$, order of auto regressive model is denoted by p, order of moving average model is denoted by q and degree of differencing is denoted by d.

Dickey-Fuller test can be used to check whether the given time series is stationary or not. Augmented dickey fuller tests are generalized dickey fuller tests which can accommodate ARIMA models.

If the time series is nonstationary differencing is performed to transform time series in to a stationary model. First order differencing can be expressed as below,

$$y_t = x_{t+1} - x_t \quad (3)$$

If $y_t$ isn't stationary, when d=1 then, it is needed to difference using d-1 times till it becomes stationary.

#### 2) Estimation of parameters

Parameter estimation involves choosing the right p and q which can describe the time series with the highest accuracy. These parameters can be approximately estimated by analyzing the auto correlation function (ACF) and the partial autocorrelation function (PACF). The ACF function can be helpful to determine the order of moving average q and PACF function can be used to determine the order of the auto regressive model. P value is the lag value where PACF graph crosses the upper confidence gap. Q value is the lag value where ACF graph crosses the upper confidence gap.

#### 3) Checking the residual diagnostics

Residual are leftovers after fitting an appropriate model. In most cases residuals can be regarded as the difference between the actual values and the forecasted values.

$$a_t = y_t - \hat{y}_t \quad (4)$$

Residual diagnostic is performed to check whether the model has been able to capture adequate information in the time series data. For an appropriate forecasting model residuals should be uncorrelated and should have zero mean.

### B. Deep learning

Deep learning model discussed in this study consists of two one dimensional convolutional layers and two LSTM layers. One dimensional convolutional layers are used to obtain lateral features from the time series data. LSTM layers are used to extract temporal features from the temperature time series. High level architecture of the deep learning model is shown below.

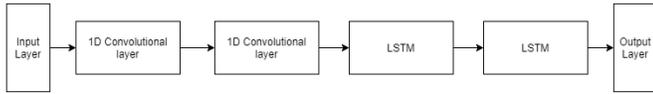

Figure.4 High level architecture of the deep learning model

1) One dimensional Convolution layer

CNNs (Convolutional neural networks) have evolved in recent years. Convolution operation is a function which is derived from two given functions where it can be used express how the shape of one function can be modified by the other function. CNNS are trained using back propagation algorithm [10]. Furthermore, CNNs are commonly used for image processing related tasks. CNNs share similar properties and similar approach without considering the dimensionality. But the main difference is from the input data dimensionality and the way filter slides across the given data. In a one dimensional convolutional layer filter slides alone a single dimension. Our temperature time series data set has two dimensions they're time steps and the temperature values. Therefore, filter or kernel can only move along the time dimension.

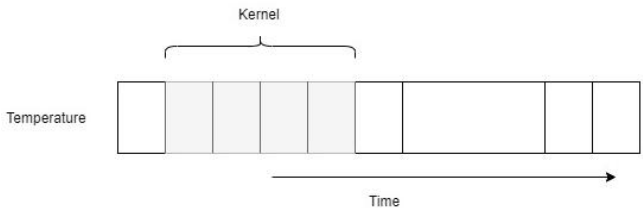

Figure.5 Kernel sliding over temperature data

Generally, pooling is applied to the filtered resulting image. The purpose of pooling is to achieve local translation invariance. Pooling achieves this purpose by down sampling the feature maps. There are two pooling methods. They're average pooling and max pooling. Initially feature maps are divided in to chunks. In max pooling maximum value of a group is taken. In average pooling average value of the group is taken. The resulting image is a shrunken version of the original.

2) LSTM Layer

According to a research paper by Hochreiter and Schmidhuber LSTMs were first introduced to the world [11]. Long short term memory networks are a special type of RNN (Recurrent neural networks). LSTMS has the ability of learning from long term dependencies. RNNs use truncated back propagation through time for training. But, RNNs suffer from vanishing gradient problem when it has too many time steps. LSTMS are designed to overcome vanishing gradient problem. LSTMs update its cell state and various gates.

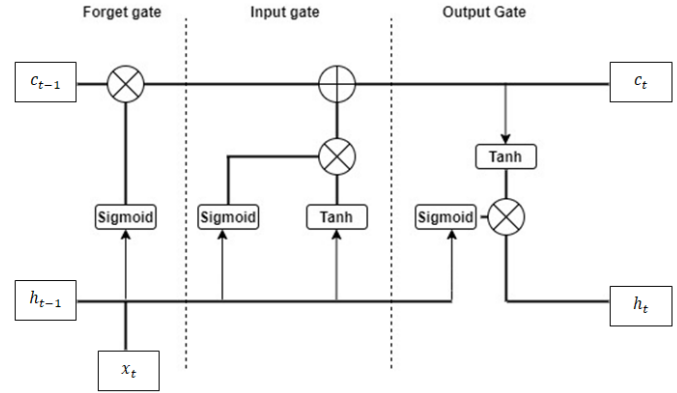

Figure.6 LSTM cell structure

Forget gate: Forget gates purpose is to decide which information should forget from the memory. For that sigmoid function is applied to information from the current input and the previous hidden state. If the value is close to zero means abandon and closer to 1 means retain.

Input Gate: Input gate concerned with updating the cell state. It first decides which values should be updated by applying sigmoid function to current input and the previous hidden state. Where 0 no updates necessary and 1 means should be updated. Then a tanh gate is applied to transform values between -1 and +1. Next outputs from these two gates are multiplied and added to the cell state.

Output gate: Output gate involves deciding the next hidden state. A sigmoid function is applied to the current input and the previous hidden state to decide which information should contain in the next hidden state. After that tanh gate is applied to the new cell state. Next hidden state is calculated multiplying these two gate values.

V. EVALUATION

To evaluate the proposed ARIMA (p,d,q) and Deep learning models same monthly temperature time series data is used for identifying ARIMA model, for training the deep learning model and testing the both of these models. For testing the both of these models datasets last 12 months monthly temperature values were used. MSE (Mean Squared Error) is used to validate the accuracy of these models in terms of mean squared error associated with each model. Lower the MSE the better the model is.

$$MSE = E[Y - \hat{Y}\ ] \qquad (5)$$

Where, $Y$ is the actual temperature and $\hat{Y}$ is the corresponding forecasted value.

## VI. RESULTS

The temperature prediction results for the last 12 months from 2016 January to 2017 January are shown below. Blue line represents the ground truth temperature value for the particular month while red line represents the prediction results for the corresponding month.

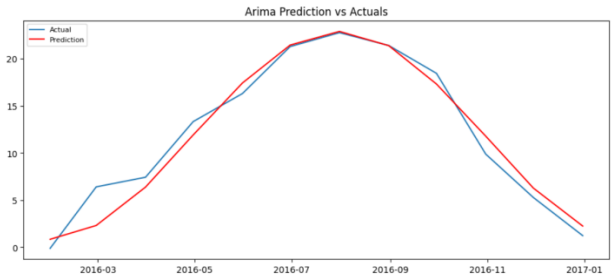

Figure.7 Forecasted results by the ARIMA model

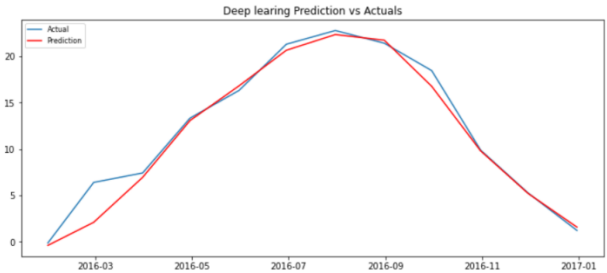

Figure.8 Forecasted results by the Deep learning model

According to figure and figure it can be clearly seen that the both models has struggle to predict the temperature at March of 2016. It can be seen as an outlier value in comparison with the March monthly temperature for other years. Furthermore, according to the both of these figures after April of 2016 deep learning model has able to forecast closer to the ground truth when compared with the ARIMA model for that given period. This shows the superiority of the deep learning model in terms of forecasting performance.

Table1. Comparison of MSE between ARIMA Model and Deep Learning Model

| Model | Mean Squared Error (MSE) |
|---|---|
| ARIMA | 2.4214 |
| Deep Learning | 1.9006 |

The numerical results on Table 1 describe the MSE comparative analysis of ARIMA and Deep learning models. According to the results it can be clearly seen that Deep learning model is superior performance in terms of accuracy than the ARIMA model as it was able to obtain the lowest mean squared error value. The MSE obtained by the deep learning model is 21% lower than the ARIMA model. Therefore, Deep leaning model has performed better than the ARIMA model.

## VII. CONCLUSION

This study has introduced Deep Learning and ARIMA models to predict temperature twelve months ahead. Both of these models were comparatively analyzed based on the performance in terms of accuracy. This paper highlights and discusses related researches. The study has used real weather data set in Szeged, Hungry for model identification and validation. According to the experimental results the deep learning model consisting of convolutional and LSTM layers outperformed the ARIMA model when forecasting temperature. In conclusion deep learning models can effectively replace traditional ARIMA models for weather parameter forecasting applications such as temperature.